\documentstyle[cogsci94]{article}

\newcommand{\entry}{\smallskip\noindent\hangindent=0.15in\hangafter=1}
\newcommand{\text}[2]{\bigskip\noindent\hangindent=55pt\hangafter=1
\hspace{15pt}\makebox[40pt][l]{\bf {#1}}\noindent{#2}}

\input psfig
\begin{document}

\title{\begin{small} Cogsci-92: In Proceedings of the Fourteenth Annual 
Conference of the Cognitive Science Society, July-August, 1992,
Bloomington, IN, pp 195-200. \end{small} \\  A Unified Process Model
of Syntactic and Semantic Error Recovery in Sentence Understanding}

\author{Jennifer K. Holbrook \\
{\normalsize Department of Psychology} \\
{\normalsize Albion College} \\
{\normalsize Albion, Michigan 49224} \\
{\normalsize jholbrook@albion.bitnet} \And
Kurt P. Eiselt \\
{\normalsize College of Computing} \\
{\normalsize Georgia Institute of Technology} \\
{\normalsize Atlanta, Georgia 30332-0280} \\
{\normalsize eiselt@cc.gatech.edu} \And
Kavi Mahesh \\
{\normalsize College of Computing} \\
{\normalsize Georgia Institute of Technology} \\
{\normalsize Atlanta, Georgia 30332-0280} \\
{\normalsize mahesh@cc.gatech.edu}}

\maketitle

\begin{abstract}
The development of models of human sentence processing has 
traditionally followed one of two paths.  Either the model posited a
sequence of processing modules, each with its own task-specific
knowledge (e.g., syntax and semantics), or it posited a single
processor utilizing different types of knowledge inextricably
integrated into a monolithic knowledge base.  Our previous work in
modeling the sentence processor resulted in a model in which different
processing modules used separate knowledge sources but operated in
parallel to arrive at the interpretation of a sentence.  One highlight
of this model is that it offered an explanation of how the sentence
processor might recover from an error in choosing the meaning of an
ambiguous word: the semantic processor briefly pursued the different
interpretations associated with the different meanings of the word in
question until additional text confirmed one of them, or until
processing limitations were exceeded.  Errors in syntactic ambiguity
resolution were assumed to be handled in some other way by a separate
syntactic module.

Recent experimental work by Laurie Stowe strongly suggests that the
human sentence processor deals with syntactic error recovery using a
mechanism very much like that proposed by our model of semantic error
recovery.  Another way to interpret Stowe's finding that two
significantly different kinds of errors are handled in the same way is
this: the human sentence processor consists of a single unified
processing module utilizing multiple independent knowledge sources in
parallel.  A sentence processor built upon this architecture should at
times exhibit behavior associated with modular approaches, and at
other times act like an integrated system.  In this paper we explore
some of these ideas via a prototype computational model of sentence
processing called COMPERE, and propose a set of psychological
experiments for testing our theories.
\end{abstract}

\section{Overview}

Most models of human language processing enforce a separation of
language levels either through an assumption of individual modules
each devoted to a different level of language or, de facto, by
focusing on only one aspect of language processing (e.g., lexical
disambiguation, theta-role assignment, or syntactic structure
building).  In contrast, our ongoing research has focused on finding
ways to integrate language processing using as few assumptions of
separate processes as possible.  However, we have always been
cognizant of the fact that theories of modular processes have support
in the literature, and we have found it convenient in our own work to
focus on lexical and pragmatic disambiguation during sentence
processing in a modular fashion.

Our current work represents a meeting of theoretical intent with
computational instantiation.  In this new model, a unified processor
is able to generate multiple inferences and make decisions among these
inferences at all levels of language processing.  Currently, our model
encompasses lexical, syntactic, semantic, and pragmatic processes.
The model is also able to make the kinds of inferential errors that
people do and to recover from them automatically, as people do.
Finally, this model, although a single processor, unites two schools
of thought regarding the modularity of language.  Our model is able to
exhibit seemingly modular processing behavior that matches the results
of experiments showing different levels of language processing (e.g.,
Forster, 1979; Frazier, 1987) but is also able to display seemingly
integrated processing behavior that matches the results of experiments
showing semantic influences on syntactic structure assignment (e.g.,
Crain \& Steedman, 1985; Tyler \& Marslen-Wilson, 1977).

\section{Background}

ATLAST (Eiselt, 1989) was a model of unified lexical and pragmatic
disambiguation and error recovery.  The model included lexical and
world knowledge; it also included some amount of syntactic knowledge.
The syntactic information was processed separately, using an ATN
parser.  The model achieved disambiguation using multiple access of
meanings for lexical items and pragmatic situations, choosing the
meaning that matched previous context, and deactivating but retaining
all other meanings.  If later context proved the initial
disambiguation decision incorrect, the retained meanings could be
reactivated without reaccessing the lexicon or world knowledge.
ATLAST proved to have great psychological validity for lexical and
pragmatic processing---its use of multiple access was well grounded in
psychological literature (e.g., Tanenhaus, Leiman, \& Seidenberg,
1979), and, more importantly, it made psychological predictions about
the retention of unselected meanings that were experimentally
validated (Eiselt \& Holbrook, 1991; Holbrook, 1989).

ATLAST was not intended to model syntactic disambiguation and error
recovery, but we believed that the principles embodied in the model
should extend to syntactic knowledge as well (Granger, Eiselt, \&
Holbrook, 1984): that syntactic disambiguation and error recovery
would follow the same pattern of multiple access, selection based on
previous context, deactivation and retention of unselected structures,
and reactivation of unselected structures should an error be
discovered.  At last year's meeting of the Cognitive Science Society,
Stowe presented the finding that syntactic information and semantic
information interact as the knowledge structure is built.  Stowe's
work (1991; Holmes, Stowe \& Cupples, 1989) has lent credence to the
prediction that syntactic knowledge is processed just like other
language knowledge sources.  Particularly relevant to the work
presented herein is Stowe's conclusion that in cases of syntactic
ambiguity, the sentence processor accesses all possible syntactic
structures simultaneously and, if the structure preferred for
syntactic reasons conflicts with the structure favored by the current
semantic bias, the competing structures are maintained and the
decision is delayed.  Furthermore, the work suggests an interaction of
the various knowledge types, as in some cases semantic information
influences structure assignment or triggers reactivation of unselected
structures.  The new psychological evidence inspired us to extend
ATLAST to include syntactic knowledge as an integral part of a unified
language processor.
  
\section{The New Theoretical Model}

We propose that the human sentence processor can best be described as
a single unified language processor which operates on distinct
knowledge sources.  These knowledge sources correspond to what are
typically labeled {\it syntax} and {\it semantics}.  While these
sources contain different types of knowledge, the same process is used
to manipulate and integrate each type of information into a coherent
and plausible interpretation.  The single processor allows inferences
about the interpretation to be generated uniformly, regardless of the
type of inference that must be made.  Thus, an ambiguous word, an
ambiguous parse tree, an ambiguous thematic role assignment, and an
ambiguous semantic representation are all disambiguated by the
processor in the same way.

This model of sentence processing attempts to explain several
different phenomena.  For example, lexical/semantic disambiguation and
error recovery are accounted for by the approach first postulated in
the ATLAST sentence processing model.  Since we are using a single
processor for all processing in our new model, the approach used by
ATLAST is now applied to syntactic disambiguation and error recovery.
As a result, we have a plausible process account of Stowe's (1991)
findings.

Additionally, because the knowledge sources are modular while the
processing is unified, we predict that this new model will sometimes
exhibit ambiguity resolution behavior like that expected from a strong
modular, autonomous process approach to sentence processing (e.g.,
Forster, 1979), and at other times it will exhibit behavior more like
that expected from a strong interactive approach (e.g., Tyler \&
Marslen-Wilson, 1977).  These differences in behavior will depend on
whether the information available from the different knowledge sources
is sufficient to resolve the specific ambiguities at hand at any given
time.  In short, this model should account for the wide range of data
accumulated by the opposing camps in this ongoing debate.

\section{Implementation}

To explore how well ATLAST's approach to lexical/semantic
disambiguation and error recovery would actually work when applied to
the resolution of syntactic ambiguity, we constructed a prototype
computational model called COMPERE (Cognitive Model of Parsing and
Error Recovery) to serve as a testbed.  This computational model
follows closely the spirit of the theoretical model described above,
but diverges slightly in actual implementation.  The divergence
appears in the processor itself: the theoretical model has a single
processor, while the prototype computational model has two
nearly-identical processors---one for syntax and one for
semantics---which share identical control structures but are
duplicated for convenience because each processor must work with
information encoded in slightly different formats.  Because we
intended only to explore syntactic ambiguity and error recovery with
this initial computational model, the distinction between two
identical processors and one unified processor is unimportant.  As we
expand the scope of our investigations, however, we will need to unify
the two processing components completely.

Both types of knowledge are represented as networks of structures.
Syntactic knowledge is represented as a network in which each node
holds all the knowledge about a particular syntactic category
necessary for parsing a sentence into its surface structure.  A node
in the network representing semantic knowledge stands for a concept.
The structure of the node (i.e., its slots) represents the
relationships between the node and other concepts.  These
relationships can include world knowledge in the form of selectional
restrictions.

In addition to syntactic and semantic knowledge, COMPERE has a lexicon
which provides the syntactic categories and subcategories of words as
well as the meanings of words represented as pointers to nodes in the
semantic network.  The semantic component also has knowledge of
thematic roles which helps bridge syntax and semantics.  For instance,
it knows that a noun phrase has a primitive role called {\bf THING}
which can evolve {\it in context} to an {\bf ACTOR} or an {\bf OBJECT}
role.  The representations of the different bodies of knowledge and
the flow of information between them is shown in Figure~1.

\begin{figure}[thb]
\ \psfig{figure=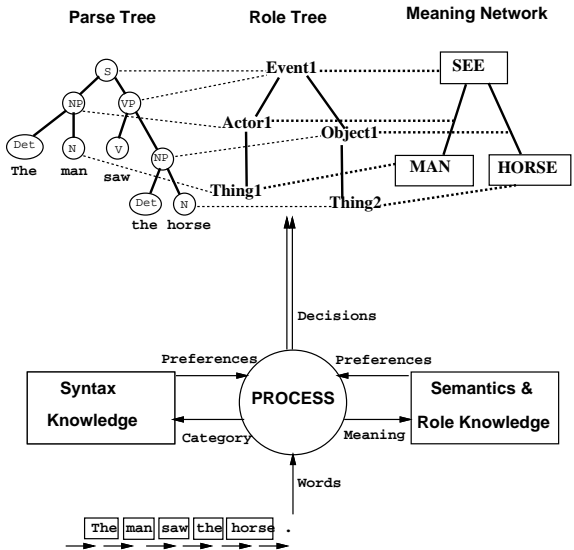,width=3.0in}
\caption{Architecture of COMPERE}
\end{figure}

\subsection{The Process}

Words are read from left to right, and their lexical entries are
retrieved.  The syntactic categories are passed to the syntax
processor; at the same time, the pointers to corresponding meanings
are passed to the semantic processor.  The semantic processor builds a
tree of thematic roles, as well as a network of instances of meaning
structures.

As explained above, the control structures of the syntax and semantic
processors are identical, though they process different kinds of
knowledge. The processors interact many times in processing each word
as they build the trees.  The syntax processor first builds the basic
node for the category of the word which will be a leaf of the parse
tree. The semantic processor builds a node for the primitive role the
word plays (if any) and also instantiates the meaning structure for
the word.  For instance, on reading the verb ``saw,'' the syntax
processor builds a verb node (V) to be added to the parse tree of the
current sentence.  The semantic processor builds nodes for an {\bf
EVENT} role and an instance of the {\bf SEE} structure. These
structures must be connected to other role and meaning structures
already built for the sentence.  The processors now try to connect the
new nodes with the partial trees built earlier. When the syntactic
structure of a sentence is successfully parsed, the meaning of the
sentence is available as the meaning attached to the root node (S) of
the parse tree.

Whenever the syntactic processor connects a node to its parent, it
communicates with the semantic processor. The semantic processor tries
to find corresponding relationships in the meanings associated with
the two nodes by way of connecting their roles in the role tree. Thus
the meanings associated with the nodes move up along the syntactic
structure. When they meet at a common node, the semantic processor
tries to bind them together through their roles.  For example,
consider the following sentence:

\text{Text 1:}{The man saw the horse.}
\medskip

\noindent The structures that exist after 
reading ``The man saw'' are shown in Figure~2.

\begin{figure}[thb]
\ \psfig{figure=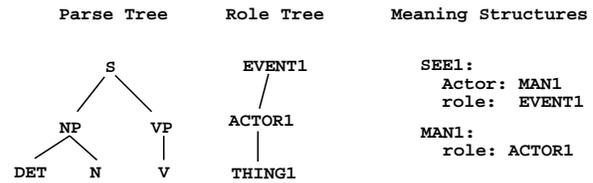,width=3.0in}
\caption{COMPERE's output for ``The man saw.''}
\end{figure}

Now, after reading ``the horse,'' the system creates a noun phrase
(NP) node to be connected to the above parse tree, a {\bf THING} role
to be connected to the above role tree, and a {\bf HORSE1} structure
to be connected to the meaning structures above. Syntactic processing
could propose a connection from the new NP to the verb phrase (VP) in
the tree, making ``the horse'' the syntactic object.  The semantic
processor finds corresponding links between the {\bf HORSE1} node and
the {\bf SEE1} node through its {\bf OBJECT} slot.  This results from
specializing the {\bf THING} role of ``the horse'' to an {\bf OBJECT}
role which can now be connected to the {\bf EVENT1} role.  This
process can be viewed as the meaning of ``horse'' propagating up the
parse tree to meet the meaning of ``see'' at the VP node where the
corresponding semantic connections are found. The structures built at
the end of the sentence are shown in Figure~3.

\begin{figure}[htb]
\ \psfig{figure=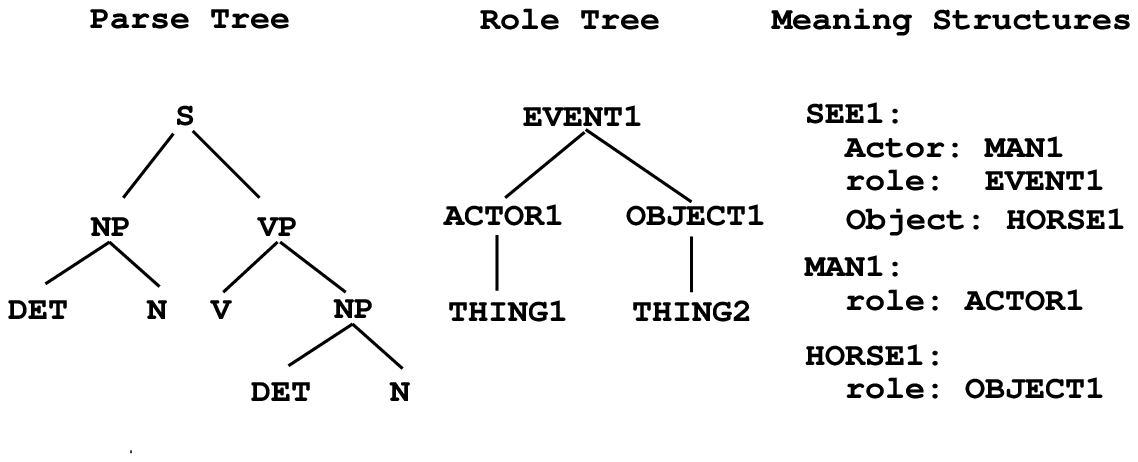,width=3.0in}
\caption{COMPERE's output for Text~1.}
\end{figure}

Though the syntactic and semantic processors interact with each other,
they are functionally independent; each can do its job should the
other fail. If the syntactic processor fails to build a parse
structure for a sentence, the semantic processor connects the
primitive role for a word with the role tree (or a set of subtrees)
built thus far. The processor can make decisions based on preferences
coming only from one source of knowledge (such as syntax or semantics)
if other sources fail to provide any preferences.  Such a failure of
the other sources could be either due to a lack of knowledge or due to
a lesion in the communication pathways.  Functionally independent
behavior of this kind would not have been possible if the system had a
single integrated source of knowledge together with a unified
processor as seen in other models (e.g., Jurafsky, 1991; Lebowitz,
1983).

\subsection{Ambiguity Resolution}

Structural ambiguities in a sentence can be resolved through semantic
or syntactic processing.  For instance, if Text~1 were changed to:

\text{Text 2:}{The man saw the woman with the horse.}
\medskip

\noindent there would be at least two possible interpretations from a 
syntactic point of view---attaching the prepositional phrase (PP) to
the VP or to the object NP---but only one of them is supported by
semantics. The NP-attachment interpretation with its ``woman together
with the horse'' meaning is acceptable whereas the VP-attachment
interpretation with its ``saw using the horse as an instrument'' is
not acceptable since it violates the constraint that the {\bf
INSTRUMENT} slot of the event {\bf SEE} must be filled by an optical
instrument.

On the other hand, consider the following sentence:

\text{Text 3:}{The officers taught at the military academy were very demanding.}
\medskip

\noindent The verb ``taught'' 
is interpreted as the main verb of the sentence since that would
satisfy the expectation of a VP at that point in processing. In other
words, we would rather use the verb to begin the VP that is required
to complete the sentence structure, instead of treating it as the verb
in a reduced relative clause which would have left the expectation of
a VP unsatisfied.  This behavior is the same as the one explained by
the ``first analysis'' models of Frazier and colleagues (Frazier,
1987) using a minimal-attachment preference.

\subsection{Error Recovery}

When choices are made to resolve structural ambiguities, the
alternatives that were not selected are retained for possible recovery
from erroneous decisions.  When it is not possible to attach a
structure to the existing tree(s), the previously retained
alternatives are examined to see if choosing another alternative at an
earlier point provides a way to attach the current structure. If so,
the tree is repaired accordingly to recover from the error. Since the
subtree that was originally misplaced is merely attached at a
different point, error recovery does not amount to reprocessing the
structure of the phrase that corresponds to the subtree.

In Text~3, until seeing the word ``were,'' the verb ``taught'' is
treated as the main verb since it satisfies the expectation of a VP
that is required to complete the sentence.  However, at this point,
the structure is incompatible with the remaining input.  The processor
now tries the other way of attaching the VP as a reduced relative
clause so that there will still be a place for a main verb. In doing
so, it did not have to process the PP that was part of the VP for the
verb ``taught.''

In resolving the structural ambiguity in Text~3, semantic preferences
did not play a significant role. In other situations, semantic
preferences could influence the decisions that the processor makes in
resolving syntactic ambiguities. Such behavior would be the same as
the ones explained by models which argue for the early effects of
semantic and contextual information in syntactic processing (e.g.,
Crain \& Steedman, 1985; Tyler \& Marslen-Wilson, 1977).  COMPERE is
intended to demonstrate that the range of behaviors that these models
account for, and the behaviors that the ``first analysis'' models
(e.g., Frazier, 1987) account for, can be explained by a unified model
with a single processor operating on multiple independent sources of
knowledge.

COMPERE has been implemented on a Symbolics workstation in the Common
Lisp language with the Common Lisp Object System. It can process both
the syntax and semantics of simple sentences (including all examples
used in this paper) and uses semantic information in resolving
structural ambiguity.  Recovery from errors in resolving structural
ambiguity has been implemented in the syntax processor alone; recovery
from lexical/semantic errors has not yet been implemented in this
model, but it will require very little effort to adapt the mechanism
already used successfully by the ATLAST (Eiselt, 1989) system.

\section{Proposed Psychological Studies}

To test the validity of our psychological claims, we must answer the
following questions: (1) How do we show that there is a single
processing architecture which applies to multiple knowledge sources to
make language decision, as opposed to multiple, non-identical
processors?  (2) How can we show error recovery occurring
automatically and on-line for lexical, syntactic, semantic and other
types of errors?

\subsection{Answering Question 1}
Recent experiments (e.g., Holmes, Stowe, \& Cupples, 1989) have
focussed on manipulating the information processed, but not the act of
processing itself.  By varying the type of task assigned to the
subject, we can manipulate the processing style that is being
executed.  We have created materials that make processing more (or
less) syntactic or semantic, by giving a task that biases the
processor toward any given level.  In one experiment, we are using two
sets of materials, one semantically weighted and the other
syntactically weighted.  We have manipulated the level of processing
by changing the task that subjects must perform.  We are comparing the
time it takes for subjects to make word-by-word completion decisions:
either a decision on whether a sentence can still be completed
grammatically, or whether a sentence can still be completed
semantically.  We are looking at the kinds of comprehension errors
that are made for syntactically versus semantically weighted
sentences, as well as at how the reaction time curve changes for the
stimuli depending on the level of processing.  Thus, in this
experiment, we are able to assess the separate effects of the
processor and the type of information processed on parsing
decisions. Both processing models make empirical predictions.  The
single-processor model predicts uniform processing errors when we
manipulate the processing environment but not the information
processed.  The multiple processor model predicts that processing
errors will be different when we manipulate the processing
environment.

A second point of comparison between single and multiple processor
models is that the single processor model assumes interaction between
lexical information and syntax and semantics, while the multiple
processor model assumes that these would be separable.  One point at
which the information sources may interact is when lexical items are
recognized.  Some words are syntactically ambiguous, such that more
than one part of speech (and probably meaning as well) must be called
up.

Seidenberg, Tanenhaus, Leiman, and Bienkowski (1982) looked at
ambiguous words that each had a meaning which lexically subcategorized
as a noun and a meaning which lexically subcategorized as a verb.
Their results showed that even when subcategorization information is
available, it does not immediately restrict the processor from viewing
all possible meanings of a word any more than other aspects of the
word's meaning do.  This is evidence that, at the place where meaning
and structure are first constructed, the information is extracted in
the same manner.  We are conducting a similar experiment to that of
Seidenberg et al., the main difference being that in our study, the
ambiguous word is embedded within the sentence instead of at the end.
This is because active suppression of alternate meanings is more
likely to occur at the end of materials than within them (Holbrook,
1989).  Seidenberg et al.'s results suggest support for the single
processor model over the multiple processor model, but only at 0 msec.
We are testing to see the time course of disambiguation due to
subcategorization information, and the extent to which
subcategorization information is relied upon exclusively for
disambiguation.  If the single processor model is correct, the
subcategorization information should be useful but not always
deterministic.  A multiple processor model would predict that the
subcategorization information will be an early and unassailable
determiner of meaning choice.

\subsection{Answering Question 2}
Error recovery ought to act differently for a single processor system
than for a multiple processor system.  A unified process ought to make
the task easy, and multiple processes ought to make it hard.  The
single processor model predicts that error recovery is uniform, no
matter at what level of processing the error occurs.  The same
elements will be brought to bear to fix the error at the lexical,
syntactic, and semantic levels.  Our previous experiments (e.g.,
Eiselt \& Holbrook, 1991; Holbrook, 1989) have validated the mechanism
for lexical ambiguity, but have not validated it for other types of
errors.  Evidence from similar experiments by Holmes, Stowe and
Cupples (1989) showed similar findings for syntactic
subcategorization: as in our experiments, one interpretation was
chosen and then discarded when later information negated this
decision.  To tie these two sets of experiments together, we are
running the variations on the Holmes et al.\ experiments described
above.  To look at error recovery, we will look for priming effects
for both meanings of the ambiguous word and for evidence of
re-instantiation of a discarded structure.

\section{Conclusion}

A model that unifies separate processing mechanisms can only be
considered successful if it is able to explain apparently different
types of output, such as syntactic and semantic output.  In this paper
we have developed a model that is able to do so by uniformly
processing different types of information.  The advantages to this
model are that processing errors are usually avoided; many of the
processing errors that still occur can be corrected immediately and
unconsciously, so that processing can remain automatic and
unconscious.  The emphasis on different information types allows our
model to remain consistent with work that suggests modularity at
various levels of processing; the modularity lies in the division of
the information types.  However, the single processor simplifies the
task of building compatible syntactic and semantic structures and
allows for their interaction as the meaning of the text is evolved
from the separate types of information.  Hence, we can explain
apparently anomalous psychological findings (e.g, Frazier, 1987; Tyler
\& Marslen-Wilson, 1977) within a single perspective.

\section{References}

\entry Crain, S., and Steedman, M. 1985.  
On not being led up the garden path:  The use of 
context by the psychological syntax processor.  
In D.R. Dowty,  L. Kartunnen, 
and A.M. Zwicky (Eds.), 
{\it Natural language parsing: Psychological,
computational, and theoretical perspectives} (pp.~320-358).
Cambridge, England:  Cambridge University Press.

\entry Eiselt, K.P. 1989.
{\it Inference processing and error recovery in sentence understanding}
(Technical Report 89-24).  Doctoral dissertation.
Irvine:  University of California, Department of Information and 
Computer Science.

\entry Eiselt, K.P., and Holbrook, J.K. 1991.  
Toward a unified theory of lexical error recovery.   
{\it Proceedings of the Thirteenth Annual Conference of the Cognitive 
Science Society}, 239-244.  
Hillsdale, NJ:  Lawrence Erlbaum.

\entry Forster, K.I. 1979.  
Levels of processing and the structure of the language processor.  
In W.E. Cooper and E.C.T. Walker (Eds.), 
{\it Sentence processing: Psycholinguistic studies presented to 
Merrill Garrett} (pp.~27-85).
Hillsdale, NJ:  Lawrence Erlbaum.

\entry Frazier, L. 1987.
Theories of sentence processing.
In J.L. Garfield (Ed.),
{\it Modularity in Knowledge Representation and Natural-Language
Understanding} (pp.~291-307).
Cambridge, MA: MIT Press.

\entry Granger, R.H., Eiselt, K.P., and Holbrook, J.K. 1984.
The parallel organization of lexical, syntactic, and pragmatic
inference processes.  {\it Proceedings of the First Annual Workshop on
Theoretical Issues in Conceptual Information Processing}, 97-106.

\entry Holbrook, J.K. 1989.
{\it Studies of inference retention in lexical ambiguity resolution}.
Unpublished doctoral dissertation.  Irvine: University of California,
School of Social Sciences.

\entry Holmes, V.M., Stowe, L.A., and Cupples, L. 1989.  
Lexical expectations in parsing complement-verb sentences.  
{\it Journal of Memory and Language}, {\it 28}, 668-689.

\entry Jurafsky, D. 1991.
An on-line model of human sentence interpretation.
{\it Proceedings of the Thirteenth Annual Conference of the Cognitive 
Science Society}, 449-454.  
Hillsdale, NJ:  Lawrence Erlbaum.

\entry Lebowitz, M. 1983.
Memory-based parsing.
{\it Artificial Intelligence}, {\it 21}, 363-404.

\entry Seidenberg, M.S., Tanenhaus, M.K., Leiman, J.M., 
and Bienkowski, M. 1982. 
Automatic access of the meanings of ambiguous words in context: 
Some limitations of knowledge-based processing.  
{\it Cognitive Psychology}, {\it 14}, 489-537.

\entry Stowe, L.A. 1991.  
Ambiguity resolution:  Behavioral evidence for a delay.  
{\it Proceedings of the Thirteenth Annual Conference of the Cognitive 
Science Society}, 257-262.  
Hillsdale, NJ:  Lawrence Erlbaum.

\entry Tanenhaus, M., Leiman, J., and Seidenberg, M.  1979.
Evidence for multiple stages in processing of ambiguous words in
syntactic contexts.  {\it Journal of Verbal Learning and Verbal
Behavior}, {\it 18}, 427-440.

\entry Tyler, L.K., and Marslen-Wilson, W.D. 1977.  
The on-line effects of semantic context on syntactic processing.  {\it
Journal of Verbal Learning and Verbal Behavior}, {\it 16}, 683-692.

\end{document}